\documentclass[10pt,conference]{IEEEtran}

\usepackage{wrapfig}
\usepackage{graphicx}
\usepackage{url}
\usepackage{booktabs}
\usepackage[table]{xcolor}
\usepackage{enumitem}
\usepackage{array}
\usepackage{tabularx}
\usepackage{multirow}
\usepackage{cite}
\usepackage{hyperref}
\usepackage{comment}               
\usepackage[T1]{fontenc}
\usepackage{tikz}
\usetikzlibrary{positioning,calc,fit,backgrounds,arrows.meta}
\def\BibTeX{{\rm B\kern-.05em{\sc i\kern-.025em b}\kern-.08em
    T\kern-.1667em\lower.7ex\hbox{E}\kern-.125emX}}

\let\oldcite\cite
\renewcommand{\cite}{\,\oldcite}
\newlist{questions}{enumerate}{2}
\setlist[questions,1]{label=\textbf{RQ$_\text{\arabic*.}$},ref=RQ$_\text{\arabic*}$, leftmargin=*}
\setlist[questions,2]{label=\textbf{1.\arabic*},ref=\thequestionsi(\arabic*), leftmargin=0cm}
\newcommand{\ques}[1]{\texorpdfstring{RQ$_{#1}$}{RQ #1}}

\definecolor{stepblue}{HTML}{8796B0}
\definecolor{candgreen}{HTML}{86B386}
\definecolor{relpink}{HTML}{C5A3B4}
\definecolor{bubblegray}{HTML}{D9D9D9}
\definecolor{bubbletext}{HTML}{5A5A5A}
\definecolor{resultpink}{HTML}{9C5C7E}

\definecolor{lavendergrey}{HTML}{8497B0}
\definecolor{powerblush}{HTML}{F7A399}
\definecolor{teagreen}{HTML}{B4C77F}
\definecolor{lightteagreen}{HTML}{DEE6C6}
\definecolor{darkteagreen}{HTML}{859C44}
\definecolor{vintageberry}{HTML}{883955}
\newcommand{\incr}[1]{\textcolor{lavendergrey}{#1}}
\newcommand{\excr}[1]{\textcolor{powerblush}{#1}}

\newcommand{\experiencepriv}[0]{$\mathit{exper}_{\mathrm{priv}}$}
\newcommand{\experiencepos}[0]{\textit{exper$_{\text{pos}}$}}
\newcommand{\experienceneg}[0]{\textit{exper$_{\text{neg}}$}}
\newcommand{\expectpos}[0]{\textit{expect$_{\text{pos}}$}}
\newcommand{\expectneg}[0]{\textit{expect$_{\text{neg}}$}}
\newcommand{\notsignificant}[0]{\textcolor{gray}{n.s.}}
\newcommand{\insight}[1]{#1}

\usepackage{makecell}

\usepackage[most]{tcolorbox}

\newtcolorbox{mybox}[1]{
  enhanced,
  breakable,
  colback=teagreen!5!white,
  colframe=lightteagreen!75!black,
  coltitle=black,
  fonttitle=\bfseries,
  title=#1,
  boxsep=0.6mm,
  left=0.6mm,
  right=0.6mm,
  top=0.6mm,
  bottom=0.6mm,
  attach boxed title to top left={xshift=2mm,yshift*=-\tcboxedtitleheight/2},
  boxed title style={
    colback=white,
    colframe=white,
    boxrule=0pt,
    sharp corners
  }
}

\usepackage{multirow}
\usepackage{orcidlink}

\begin{document}
\title{How Developer Personality Shapes the Use of AI Coding Assistants}

\author{\IEEEauthorblockN{1\textsuperscript{st} Annemarie Wittig\orcidlink{0000-0002-7051-2242}}
\IEEEauthorblockA{\textit{Leipzig University}, \\ Leipzig, 04109, Germany}
\and
\IEEEauthorblockN{2\textsuperscript{nd} Alina Mailach\orcidlink{0000-0001-6204-2095}}
\IEEEauthorblockA{\textit{ScaDS.AI Dresden/Leipzig}, \\ \textit{Leipzig University}, \\ Leipzig, 04109, Germany}
\and
\IEEEauthorblockN{3\textsuperscript{rd} Norbert Siegmund\orcidlink{0000-0001-7741-7777}}
\IEEEauthorblockA{\textit{ScaDS.AI Dresden/Leipzig}, \\ \textit{Leipzig University}, \\ Leipzig, 04109, Germany}
}

\maketitle

\begin{abstract}

Developers differ in personality traits that shape how they approach software development tasks, yet AI coding assistants either behave uniformly or offer personalization features whose fit with individual developers must be found through trial and error. 
This is because it remains unclear how personality shapes developers' experiences with AI assistants, and thus which traits personalization should account for.
In other words, we provide tools with no underlying theory on what works when and why. 
We take a first step towards such a theory with a systematic scoping review on how users' personality relates to their perception of LLM-based tools. 
We identified twelve studies through general search engines, major Software Engineering (SE) venues, AI tooling, and snowballing.
They all reported associations between personality and perceptions, particularly perceived usefulness, yet none addressed this in a SE context.

Building on these findings, we conducted an exploratory two-wave survey with 
29 and 11 developers, collecting first empirical evidence of how developers' personality (measured using the Big Five Model) relates to the perception of AI coding assistants. 
Higher Extraversion was associated with more positive private-use experiences in both waves, while Agreeableness was associated with negative professional experiences in the first wave and positive ones in the second. 
These results provide first indications that the personality effects found in the literature extend to developers.
Although further studies with bigger samples are needed to build a theory, our work takes the first step towards an empirical grounded theory of developer personality and AI personalization traits.

\end{abstract}
\section{Introduction}

How we think, collaborate, and engage with the world around us is fundamentally shaped by who we are. 
In Software Engineering, this manifests in personality traits suiting different tasks and development phases\cite{personality_diversity,software_dev_and_personality}, impacting how teams function\cite{software_developer_character_in_teams}, and even affecting the produced software quality\cite{personality_traits_and_software_quality}. From tasks they take on and software they build, personality shapes how developers engage with their craft. 
Beyond how developers act, the \emph{perception} and \emph{adoption} of software engineering tools are unlikely to be exempt from this.

Among the tools developers have adopted most rapidly, LLM-based coding assistants and agents stand out. 
According to the Stackoverflow surveys, 63.2\,\% of professional developers used AI tools in 2024\cite{stackoverflow2024survey}.
By 2025, that number rose to 80.8\,\%\cite{stackoverflow2025survey}, and, according to DORA's 2025 report, even more to 90\,\% of all technology professionals\cite{dora2025}.
This adoption was accompanied by a growing body of research investigating the effects of AI coding assistants on developers.
A common topic of such analyses is developer productivity, including technology-focused outcomes, such as number of pull requests\cite{beyond_code,the_effects_of_generative_ai_on_high_skilled_work,ai_assisted_programming_may_decrease_the_productivity_of_experienced_developers_by_increasing_maintenance_burden} and lead time for changes\cite{the_ai_productivity_paradox}, but also human-centered effects, such as satisfaction\cite{harnessing_the_potential_of_gen_ai_coding_assistants_in_public_sector_software_development,significant_productivity_gains_through_programming_with_large_language_models,the_impact_of_ai_pair_programmers_on_code_quality_and_developer_satisfaction,developers_perceptions_on_the_impact_of_chatgpt_in_software_development}, stress and well-being\cite{investigating_the_impact_of_ai_assisted_tools_on_software_practitioner_well_being}. 
This body of work suggests that adoption and productivity gains are not equally distributed among developers, with differences (partly) attributed to factors such as work experience and age\cite{the_impact_of_ai_on_developer_productivity} or task type and complexity\cite{coding_assistants_in_practice,developers_perceptions_on_the_impact_of_chatgpt_in_software_development,harnessing_the_potential_of_gen_ai_coding_assistants_in_public_sector_software_development,the_space_of_ai}. 

\begin{figure}[tbp]
  \begin{center}
    \includegraphics[width=0.8\linewidth]{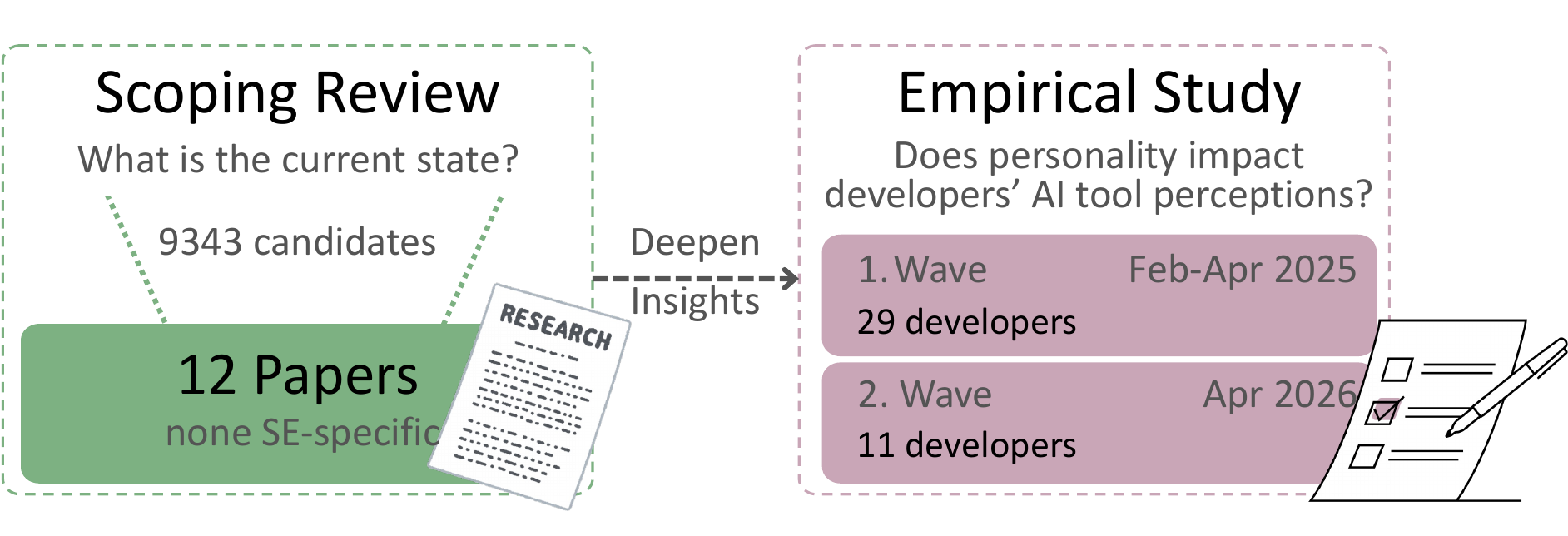}
     
    \caption{Study Structure}
    \label{fig:structure}
    \end{center}
\end{figure}

At the same time, user personalization and customization of LLM tools have emerged as an active direction in human-AI research and industry alike.
For instance, Liu et al.\cite{liu2025surveypersonalizedlargelanguage} show that various approaches to personalizing LLM-based systems to improve user experience exist. 
In their survey, they find more than 50 personalization methods, including personalized prompting, model adaptation, and alignment.
Similarly, in a recent interview, Tibo, Head of Product at OpenAI, described a new generation of personalized agents that can understand users' preferences and goals and eventually ``take the initiative to act at the right time'', rather than merely respond to explicit queries\cite{interview}. 
If this vision holds, browser-based assistants, IDE-integrated coding tools, and autonomous agents will increasingly tailor their behavior to individual users.

Despite these efforts towards personalization, we still lack key knowledge of how users' individual characteristics shape their experience with AI assistants, and thus which traits personalization should account for.
We argue this understanding should come first, since the experience depends on both the tool and its user.
Given the established role of personality in software engineering, it is surprising that, to the best of our knowledge, no prior work has examined how personality relates to the adoption and perception of coding assistants.

Instead of focusing on the person using AI assistants, the SE community rather focused on the personality of the assistant itself, such as what personalities LLMs can reliably (and unreliably) display\cite{designing_ai_agents_with_personalities,llms_self_reporting,evaluating_llm_alignment_on_personality,personallm}, which LLM-traits users prefer\cite{alexas_personality,evaluating_llm_alignment_on_personality,older_adults_personality_preferences}, and how to assign LLMs specific traits\cite{designing_ai_agents_with_personalities,big5_chat}. 
Even when the user's personality is considered, it is examined for its alignment with the LLM's personality\cite{differential_effects_of_agreeableness_and_extraversion,vibe_check} as opposed to examining how the user's personality shapes interactions, expectations, and experiences with LLM-based tools. 
This gap is remarkable, especially considering that research in other domains suggests that personality dimensions predict how users perceive LLM-based tools\cite{exploring_peoples_perceptions_of_llm_generated_advice,personality_traits_and_the_technology_acceptance_of_chatgpt}.
In essence, current approaches matching LLMs to users' personalities are developed before we know what users with different personalities actually need from these tools.

To understand this relationship systematically, we conducted a systematic literature review synthesizing empirical studies on personality and AI-tool perception in Software Engineering and beyond.
Across twelve identified studies, different personality traits predict user perceptions, particularly perceived usefulness and ease of use, though the specific associations varied between traits and outcomes. 
These studies establish personality as a relevant predictor in AI-tool perception, yet one whose effects appear context-dependent. 
Critically, no corresponding evidence exists in the SE field, which is itself an important contribution as it clearly points to a missing piece in the interplay of humans and AI assistants in SE. 
Still, it leaves the question whether such associations also manifest for developers, which we investigated in a second study.

To this end, we conducted a two-wave survey across three Software Engineering companies (29 initial, 11 follow-up responses \textasciitilde{}1 year later), relating Big Five personality traits\cite{SotoJohn2017} to developers' expectations, workplace experiences, and perceptions of AI tools in private programming. 
Our findings: Associations emerged for personality and AI-tool perceptions across both waves, particularly for developers' actual experiences rather than their expectations.
This represents the first Software Engineering-specific evidence that personality is associated with how developers perceive coding assistants, though its expression varies by context and experience.

While further research is needed to confirm and extend these patterns, our initial findings point towards personality-informed approaches to tool introduction, training, and support as future directions worth exploring.
For organizations providing AI coding tools at scale, uniform rollouts and support structures may not benefit all developers equally. Personality appeared to relate more strongly to developers' actual experiences than their expectations, and associations varied between waves. If these tendencies hold in larger samples, individual differences may become most visible during everyday use rather than at introduction. 
In that case, support that extends beyond initial onboarding and adapts to individual experiences could be more effective than one-time measures.

To summarize, our contributions are as follows:
\begin{itemize}[label={}, leftmargin=0pt]
    \item A \textbf{systematic scoping review} mapping research on associations between user personality and perceptions of LLM-based tools, establishing personality as relevant across contexts while identifying the absence of corresponding SE-specific evidence.
    \item \textbf{First SE-specific empirical evidence} that personality is associated with developer experiences with AI tools, through a two-wave survey across three SE companies relating Big Five traits to expectations and experiences with coding assistants.
    \item A \textbf{comprehensive replication package} comprising all survey data and analysis scripts, supporting reproducibility and investigation of alternative hypotheses shared on Zenodo\cite{zenodo}.
\end{itemize}

\section{Related Work}
As our study sits at the intersection of personality research and developer experiences with AI tools, we first introduce the Big Five Model of Personality, then discuss related work from both areas.

\paragraph{The Big Five Model of Personality}

Personality psychology has produced various theories on personality. 
One of the ones used most often in empirical research is the Big Five (BF) Model of Personality\cite{bfi44}. Derived from a lexical approach, it structures personality into five traits: Extraversion (sociability, assertiveness, energy level), Agreeableness (compassion, respectfulness, trust), Conscientiousness (organization, productiveness, responsibility), Negative Emotionality (anxiety, depression, emotional volatility; formerly named Neuroticism), and Open-Mindedness (intellectual curiosity, creative imagination, aesthetic sensitivity; formerly Openness)\cite{bfifull}. 
A key reason for the success of the BF Model is the large number of measurement instruments, ranging from long-form inventories to brief short forms\cite{bfi44, bfi8, SotoJohn2017}, and its factors have been shown to be predictive for a wide range of life outcomes, such as job performance, academic achievement, health, and well-being\cite{ozer2006personality, barrick1991big}. This makes the BF Model a natural candidate for studying how individual differences shape the way people engage with technology. 

\paragraph{Personality and Software Development}

In Software Engineering, personality's effect has been studied extensively, using the BF Model as well as other frameworks. A systematic mapping study identified 90 such studies across nine research topics, including pair programming, team effectiveness, and software process allocation, but also considerable variation and contradictory findings across contexts and research methods\cite{mapping_study}.
At the team level, Yilmaz et al.\cite{software_developer_character_in_teams} found that effective teams were characterized by members with certain personality traits, and in pair programming, certain personalities prefer navigator roles while others prefer pilot or solo roles\cite{personality_pair_programming}. At the individual level, personality has been shown to affect preferences for agile development approaches\cite{toward_an_understand_of_perference_for_agile_software_development_methods} and was associated with the quality of the code produced by students, though not uniformly across all measured quality metrics\cite{personality_traits_and_software_quality}.

Beyond Software Engineering, personality has also been related to general attitudes towards AI, for example using the General Attitudes towards Artificial Intelligence Scale (GAAIS)\cite{schepman2020initial}, with findings such as more positive attitudes among introverts\cite{the_general_attitudes_towards_artificial_intelligence_scale} and further trait associations in subsequent studies\cite{gaaisungarn,gaaisfinance,gaaisnursing}.
However, these studies treat AI broadly, e.g., recommender systems and autonomous vehicles, leaving unclear which systems participants had in mind and how the findings transfer to LLM-based tools.

Despite the established role of personality in Software Engineering, it remains largely absent from research on how developers interact with and experience LLM-based tools (hereafter also \emph{AI tools}). If personality shapes task preferences, team dynamics, and even code quality, it seems plausible that it similarly shapes how developers perceive and adopt AI coding assistants.
With our study, we want to take a first step towards finding such associations between developer personality and their perceptions of AI coding assistants.

\paragraph{Developer Experiences with AI}

The effects of AI tool use on software developers have been studied from a variety of perspectives, including human-centered outcomes, such as satisfaction\cite{harnessing_the_potential_of_gen_ai_coding_assistants_in_public_sector_software_development,significant_productivity_gains_through_programming_with_large_language_models,the_impact_of_ai_pair_programmers_on_code_quality_and_developer_satisfaction,developers_perceptions_on_the_impact_of_chatgpt_in_software_development}, happiness\cite{the_space_of_ai}, and well-being\cite{investigating_the_impact_of_ai_assisted_tools_on_software_practitioner_well_being}, and performance-oriented outcomes, such as task completion time\cite{the_impact_of_ai_on_developer_productivity,the_impact_of_ai_pair_programmers_on_code_quality_and_developer_satisfaction,harnessing_the_potential_of_gen_ai_coding_assistants_in_public_sector_software_development,significant_productivity_gains_through_programming_with_large_language_models}, development speed\cite{the_impact_of_ai_tools_on_software_development,my_code_is_less_secure_with_gen_ai,youre_on_a_bicycle_with_a_little_motor,practices_and_challenges_of_using_github_copilot}, and code quality\cite{significant_productivity_gains_through_programming_with_large_language_models,the_impact_of_ai_tools_on_software_development,youre_on_a_bicycle_with_a_little_motor,practices_and_challenges_of_using_github_copilot}, with developer productivity as a central focus\cite{developers_perceptions_on_the_impact_of_chatgpt_in_software_development,harnessing_the_potential_of_gen_ai_coding_assistants_in_public_sector_software_development,echoes_of_ai,the_impact_of_ai_tools_on_software_development}.

Secondary studies synthesizing this research generally suggest an overall positive effect of AI tools, while also revealing substantial variations across studies, outcomes, and contexts\cite{wittig,mohamed}.
For example, Wittig et al.\cite{wittig} found positive tendencies in reported productivity metrics but more negative direct assessments of the AI tools, and Mohamed et al.\cite{mohamed} identified eight benefits, such as accelerated development, and five risks, such as disrupted flow, with code quality appearing as both.
Wittig et al.\cite{wittig} further found that not only do effects vary, but also several context- and person-related variables likely affecting productivity assessments remain underrepresented. 
Our study takes a first step towards understanding the role of personality in developers' experiences with AI tools.

\paragraph{LLM Personality and User Perception}
Research on personality in the context of LLMs has mostly focused on the personality of the tool rather than of its user.
Prior work has assigned LLMs certain character traits, either through prompt engineering\cite{designing_ai_agents_with_personalities,personallm,differential_effects_of_agreeableness_and_extraversion,vibe_check} or fine-tuning\cite{big5_chat}, often using BF-based instruments to assign personas\cite{designing_ai_agents_with_personalities,big5_chat,differential_effects_of_agreeableness_and_extraversion}, assess an assistant's displayed personality\cite{alexas_personality}, or evaluate trait consistency\cite{llms_self_reporting}.
Such personas affect how users perceive the tool. 
One study had participants plan a trip using LLMs with low, medium, or high expressions across all Big Five traits and found the medium-expressive model rated significantly higher than the other versions\cite{vibe_check}. 
This aligns with broader evidence that certain traits are perceived as more likable, empathetic, or otherwise positive\cite{differential_effects_of_agreeableness_and_extraversion,older_adults_personality_preferences,vibe_check}.

The user's personality enters this research mainly through its (mis)match with the tool's personality.
Mathur et al.\cite{differential_effects_of_agreeableness_and_extraversion} found that older adults' perceptions of LLM-based voice assistants were only affected by mismatches in Agreeableness, while Rahman and Desai\cite{vibe_check} found that mismatches in different dimensions affected different perceptions.  
While these studies underscore the relevance of personality in human–tool interaction, they approach it from the tool side, rather than examining how the user's own personality shapes their perception. We examine this gap systematically through the following scoping review.

\section{Research Questions}

With this study, we want to address two research questions:
\begin{questions}
    \item How is the personality of LLM-based tool users associated with their perceptions of the tools?
    \item How are developers' personality traits associated with their expectations of, experiences with, and attitudes toward LLM-based coding assistants?
\end{questions}

\ques{1} addresses the question of how personality is generally associated with AI tool perceptions, within and beyond Software Engineering.
The rationale is that to provide realistic personalization in AI tools, we need to understand the base needs of users first.
To address this, we conduct a systematic scoping review to collect and identify evidence of such associations in the literature.
We then use \ques{2} to examine whether the associations identified in the review extend and manifest in the specialized Software Engineering context. 
To address this, we conducted a two-wave survey study across three companies, examining associations between Big Five personality dimensions and developers' expectations, experiences, and attitudes toward coding assistants

\section{Systematic Scoping Review}

In this section, we present our systematic scoping review that examines and collects evidence on how personality is associated with perceptions of LLM-based tools according to \ques{1}.
A scoping review systematically maps the extent, range, and nature of literature on a topic while identifying gaps\cite{scoping}.

\subsection{Methodology}
We follow established guidelines for conducting and reporting scoping and systematic literature reviews\cite{scoping,kitchenham2007guidelines}. We provide the supplementary material of this survey, containing all scripts and documentation (such as full search strings and detailed inclusion/exclusion criteria). This also includes a review protocol and a set of quality attributes for Systematic Reviews, derived from the ACM SIGSOFT Empirical Standards\cite{ralph2021empiricalstandardssoftwareengineering}, which we adhere to.

\subsubsection{Selection Criteria}

\newcolumntype{Y}{>{\raggedright\arraybackslash}X}

\begin{table}[t]
\centering
\footnotesize
\caption{Inclusion and exclusion criteria.}
\label{tab:criteria}
\begin{tabularx}{\linewidth}{@{}p{0.04cm} p{0.09cm} p{1.5cm} Y@{}}

\toprule
 & \textbf{ID} & \textbf{Aspect} & \textbf{Criterion} \\
\midrule
\multirow{13}{*}{\rotatebox[origin=c]{90}{\textit{Inclusion}}}
 & \incr{S1} & Study type & Primary data source is human-centered study. \\
 & \incr{S2} & Topic & Studies the impact of the user's personality on their perception of LLMs; personality must appear as a research question, hypothesis, or other primary research objective. \\
 & \incr{S3} & Technology & Focuses on LLMs, AI coding assistants, or general-purpose LLM-based tools; no specialized non-SE LLM-applications. \\
 & \incr{S4} & Context & Any study context. \\
 & \incr{S5} & Personality measure & Assesses personality with an established multidimensional personality framework. \\
 & \incr{S6} & Availability & Publicly available in English. \\
\midrule
\multirow{8}{*}{\rotatebox[origin=c]{90}{\textit{Exclusion}}}
 & \excr{E1} & \makecell{Tool\\personality} & Studies the impact of the AI tool's personality on participants' perceptions. \\
 & \excr{E2} & Personality match & Studies the (mis)match between AI tool and user personality without examining the impact of user personality on its own. \\
 & \excr{E3} & Publication date & Published before 2022 (before LLMs were widely adopted). \\
 & \excr{E4} & Peer review & Not peer-reviewed. \\
\bottomrule
\end{tabularx}
\end{table}

We define six inclusion and four exclusion criteria, as presented in \autoref{tab:criteria}. The selection criteria reflect different aspects of our research question.
We therefore restrict our candidate papers to studies with human participants (\incr{S1}) and require personality to be an explicitly investigated variable, for example through a research question, hypothesis, or other primary research objective (\incr{S2}). 
This ensures that the included studies provide interpretable evidence on the relationship between personality and perception rather than treating personality only as an incidental finding if at all. 
Accordingly, we exclude studies examining personality expressed by or assigned to the LLM (\excr{E1}) as well as studies concerned only with personality congruence or mismatch between user and tool (\excr{E2}), as these studies investigate related but conceptually different relationships to the one in our study.
As personality is a complex construct, we require a validated multidimensional personality framework (\incr{S5}) and exclude non-validated trait measures, even reported alongside validated ones. 
This ensures that the included personality constructs are psychometrically grounded and reasonably comparable across studies. 

We restrict the investigated systems to general-purpose LLMs and LLM-based coding assistants, excluding specialized domain-specific applications outside SE (\incr{S3}). 
For studies using broader terminology such as \textit{generative AI}, inclusion was determined by whether the paper's conceptual framing and operationalization primarily referred to LLM-based tools, rather than AI in general. 
Specialized tools, by contrast, are optimized for particular tasks and contexts, limiting transferability.
Finally, we restrict the corpus to publicly available English-language studies (\incr{S6}), peer-reviewed work (\excr{E4}), and publications from 2022 onward (\excr{E3}) to provide a consistent evidence baseline focused on the period of widespread availability and adoption of LLM-based tools.

\subsubsection{Paper Collection and Screening Procedure}

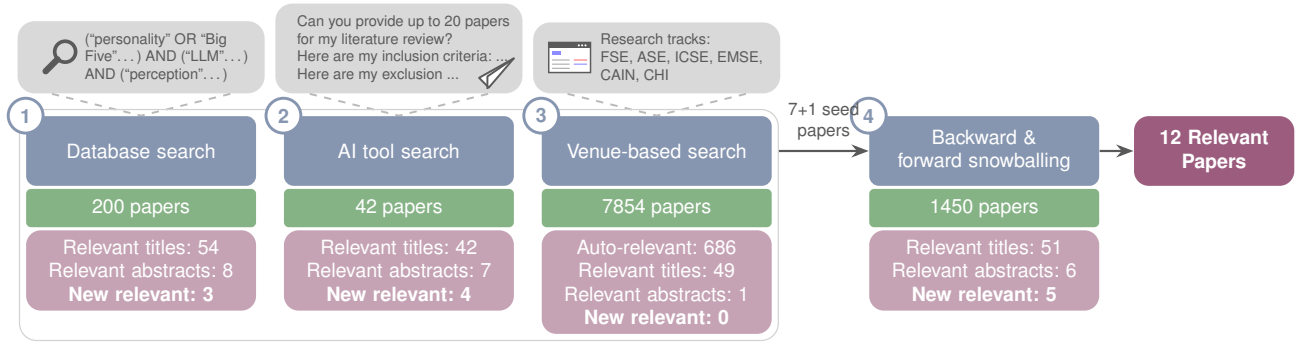
\begin{figure}
    \centering
\resizebox{1\linewidth}{!}{
\def\cw{3.3}   
\def\tw{3.0}   
\def\figS{\fontsize{8}{9.5}\selectfont}    
\def\figF{\fontsize{7}{8.5}\selectfont}    
\def\figX{\fontsize{6}{7.2}\selectfont}    
\begin{tikzpicture}[
  font=\sffamily,
  method/.style = {rounded corners=4pt, fill=stepblue, text=white, align=center,
                   text width=\tw cm, minimum width=\cw cm, minimum height=1.0cm,
                   font=\sffamily\figS},
  cand/.style   = {rounded corners=2pt, fill=candgreen, text=white,
                   font=\sffamily\figS, minimum width=\cw cm, minimum height=0.55cm},
  stats/.style  = {rounded corners=6pt, fill=relpink, text=white, align=center,
                   text width=\tw cm, minimum width=\cw cm, inner sep=4pt,
                   font=\sffamily\figS},
  btext/.style  = {text=bubbletext, align=left, font=\sffamily\figX,
                   anchor=south west, inner sep=1pt},
  bubble/.style = {rounded corners=7pt, fill=bubblegray, inner sep=3.5pt},
  callout/.style= {dashed, gray!55, line width=0.8pt},
  line/.style   = {line width=0.9pt, gray!70!black, rounded corners=4pt},
  arrow/.style  = {line, -{Stealth[length=2.5mm]}},
  badge/.style  = {circle, fill=white, draw=stepblue, line width=1.2pt,
                   text=stepblue, font=\sffamily\figS\bfseries, inner sep=1pt,
                   minimum size=0.5cm},
  result/.style = {rounded corners=6pt, fill=resultpink, text=white, align=center,
                   text width=2.0cm, minimum width=2.3cm, minimum height=1.0cm,
                   font=\sffamily\figS\bfseries},
]
  \def\colA{0}
  \def\colB{3.7}
  \def\colC{7.4}
  \def\colD{12.1}   
  \def\colR{15.4}
  \def\by{0.95}     

  \node[method] (h1) at (\colA,0) {Database search};
  \node[cand,  below=0.03cm of h1] (c1) {200 papers};
  \node[stats, below=0.03cm of c1] (r1)
    {Relevant titles: 54\\ Relevant abstracts: 8\\ \textbf{New relevant: 3}};
  \node[btext] (t1) at (\colA-0.85,\by)
    {(``personality'' OR ``Big\\Five''\ldots) AND (``LLM''\ldots)\\ AND (``perception''\ldots)};
  \coordinate (i1) at ($(t1.west)+(-0.3,0.02)$);
  \draw[bubbletext, line width=1.5pt] ($(i1)+(0.06,0.06)$) circle (0.15);
  \draw[bubbletext, line width=2.6pt, line cap=round]
        ($(i1)+(-0.05,-0.05)$) -- ($(i1)+(-0.22,-0.22)$);
  \coordinate (i1a) at ($(i1)+(-0.3,0)$);

  \node[method] (h2) at (\colB,0) {AI tool search};
  \node[cand,  below=0.03cm of h2] (c2) {42 papers};
  \node[stats, below=0.03cm of c2] (r2)
    {Relevant titles: 42\\ Relevant abstracts: 7\\ \textbf{New relevant: 4}};
  \node[btext] (t2) at (\colB-1.5,\by)
    {Can you provide up to 20 papers\\for my literature review?\\
     Here are my inclusion criteria: ...\\
     Here are my exclusion ...};
  \coordinate (i2) at (\colB+1.15,\by+0.1);
  \draw[bubbletext, line width=0.8pt, line join=round, fill=white]
        (i2) -- ++(0.5,0.28) -- ++(-0.36,-0.42) -- cycle;
  \draw[bubbletext, line width=0.8pt, line join=round]
        ($(i2)+(0.5,0.28)$) -- ($(i2)+(0.12,0.02)$) -- ($(i2)+(0.12,-0.16)$);
  \coordinate (i2a) at ($(i2)+(0.5,0.28)$);

  \node[method] (h3) at (\colC,0) {Venue-based search};
  \node[cand,  below=0.03cm of h3] (c3) {7854 papers};
  \node[stats, below=0.03cm of c3] (r3)
    {Auto-relevant: 686\\ Relevant titles: 49\\ Relevant abstracts: 1\\ \textbf{New relevant: 0}};
  \node[btext] (t3) at (\colC-0.85,\by)
    {Research tracks:\\ FSE, ASE, ICSE, EMSE,\\CAIN, CHI};
  \begin{scope}[shift={($(t3.west)+(-0.7,-0.21)$)}, scale=0.6]
    \draw[bubbletext, fill=white, line width=0.6pt] (0,0) rectangle (1,0.72);
    \fill[bubbletext] (0,0.58) rectangle (1,0.72);
    \foreach \d in {0.72,0.8,0.88} \fill[white] (\d,0.65) circle (0.025);
    \fill[blue!45] (0.08,0.43) rectangle (0.28,0.52);
    \foreach \yy in {0.3,0.2,0.1}
      \draw[gray!60, line width=0.5pt] (0.08,\yy) -- (0.5,\yy);
    \draw[red!50,  line width=0.5pt] (0.6,0.3) -- (0.92,0.3);
    \draw[blue!40, line width=0.5pt] (0.6,0.2) -- (0.92,0.2);
    \coordinate (i3)  at (0,0);
    \coordinate (i3a) at (1,0.72);
  \end{scope}

  \node[method] (h4) at (\colD,0) {Backward \& forward snowballing};
  \node[cand,  below=0.03cm of h4] (c4) {1450 papers};
  \node[stats, below=0.03cm of c4] (r4)
    {Relevant titles: 51\\ Relevant abstracts: 6\\ \textbf{New relevant: 5}};
  \node[result] (res) at (\colR,0) {12 Relevant Papers};

  \foreach \i in {1,2,3,4}
    \node[badge] (k\i) at (h\i.north west) {\i};

  \foreach \i/\x in {1/\colA, 2/\colB, 3/\colC} {
    \coordinate (L\i) at (\x-0.5*\cw,\by);
    \coordinate (R\i) at (\x+0.5*\cw,\by);
  }
  \begin{scope}[on background layer]
    \node[bubble, fit=(t1)(i1a)(L1)(R1)]      (b1) {};
    \node[bubble, fit=(t2)(i2)(i2a)(L2)(R2)]  (b2) {};
    \node[bubble, fit=(t3)(i3)(i3a)(L3)(R3)]  (b3) {};
    \foreach \i in {1,2,3} {
      \draw[callout] ($(b\i.south west)+(0.5,0)$) -- (h\i.north);
      \draw[callout] ($(b\i.south east)+(-0.5,0)$) -- (h\i.north);
    }
  \end{scope}

  \begin{scope}[on background layer]
    \node[draw=gray!35, line width=0.5pt, rounded corners=7pt,
          inner sep=2.5pt, fit=(h1)(h2)(h3)(r1)(r2)(r3)] (grp) {};
  \end{scope}
  \draw[arrow] (grp.east |- h4.west) -- (h4.west);
  \node[above=1pt, font=\sffamily\figF, text=gray!70!black, align=center]
    at ($(grp.east |- h4.west)!0.5!(h4.west)$) {7+1 seed\\papers};

  \draw[arrow] (h4.east) -- (res.west);
\end{tikzpicture}}
\caption{Scoping review structure and intermediate results}
     \label{fig:review_results}
 \end{figure}

We used two meta databases, three AI tools, and a venue-based screening approach to collect candidate papers for our literature review.
We present an overview of each step in \autoref{fig:review_results}.
While the database and AI search kept our study broad and open to various fields, directly searching leading SE-specific venues ensured their coverage.
The review was conducted by the first author, discussing any ambiguous papers with the second. 
To validate the screening, the second author independently re-screened the first 20 relevant titles of the manual search and all relevant titles from the venue-based search.
Next, we will describe each search step in detail.

\paragraph{Database search}

We built two search strings, one inspired by the PICO framework (Population, Intervention, Comparison, and Outcome)\cite{richardson1995well} and one boolean search string built by us, and combined them with three static search strings we created based on our domain knowledge. 
We adapted the PICO framework by constructing a Boolean search string combining terms representing (P) personality-related constructs (e.g., “personality”, “Big Five”, “individual differences”), (I) LLM-based tools (e.g., “LLM”, “large language model”, “coding assistant”, etc.), and (O) perception-related outcomes (e.g., “perception”, “trust”, etc.). No comparison (C) component was included, as studies in this domain rarely report explicit control conditions in titles or abstracts, if they exist at all.
The boolean string focuses more on SE-specific settings by replacing the (O) with software developer terms.
The static strings comprise SE-specific and general settings, capturing research on both AI users broadly and developers specifically.
They were queried against Google Scholar and Semantic Scholar, two general research databases.
For each executed search query, we manually reviewed the first 20 results by judging whether the corresponding titles are relevant with respect to our selection criteria. As results are ranked by relevance, we predefined a cutoff of 20 per query and broadened coverage through the subsequent search steps.
This process resulted in us reviewing \textbf{200} candidate papers.

\paragraph{AI Tool search}
We used three separate AI tools (Claude\footnote{\url{https://claude.ai/} (Sonnet 4.6)}, Google Scholar Labs\footnote{\url{https://scholar.google.com/scholar_labs/} (underlying model not disclosed)}, and Google AI Studio\footnote{\url{https://aistudio.google.com/} (underlying model not disclosed)}) to support our paper retrieval process. 
A preliminary search revealed that this procedure improved the coverage of non-SE fields. 
For each tool, a single prompt was issued on May 12th 2026 incorporating the study's inclusion and exclusion criteria, asking for up to 20 suitable papers. 
For the conversational assistants (Google AI Studio and Claude), the prompts were issued in a private chat to ensure no context interference, and their responses are documented in the supplementary materials. 
For Scholar Labs, results were directly transferred to the candidate list without separate documentation, as it does not provide conversational answers. 
The \textbf{42} recommended papers were included as candidate papers.
In the following screening step, we verified the correctness and existence of all papers and excluded hallucinations.

\paragraph{Venue-based search}
To extend coverage beyond search engine results and reduce bias from our search strings, we additionally conducted a venue-based search at selected venues, reducing the risk of missing relevant work.
Specifically, we selected ICSE, ASE, FSE, and EMSE as the leading SE venues, CAIN as a dedicated venue for SE of AI-enabled systems, and CHI to capture high-quality human-centered AI research published in a leading venue outside of SE.
For each venue, we first collected all research-track papers published between 2023 (i.e., submission in 2022) and 20th May 2026, resulting in \textbf{7854} papers.
While \excr{E3} allows for papers from 2022 onward, we focused the venue-based search on 2023–2026 due to typical publication time in major SE venues, where work initiated during the initial adoption phase of AI-assisted software engineering in 2022 is predominantly published from 2023 onward. 

To reduce manual effort, we pre-screened the candidate pool using keyword matching, mirroring manual title-screening practice: a title was flagged as potentially relevant if it contained both an LLM-related keyword (out of 17 options, e.g., \textit{LLM}, \textit{GenAI}, \textit{coding assistants}) and a human-study or personality-related keyword (out of 30 and 27 options respectively, e.g., \textit{user}, \textit{satisfaction}, \textit{personalit*}, \textit{BFI}). 
We validated the pre-screening on the SE venue papers, where all candidates from an independent manual title screening were among the automatically flagged titles.
Flagged titles then entered the standard title-screening pipeline described above.
Keywords were chosen based on domain knowledge and are provided in the supplementary material.

\paragraph{Screening}
After title screening, we assessed each paper's abstract with a relevant title against our inclusion and exclusion criteria. 
Abstracts that were ambiguous against these criteria were marked as maybe relevant. 
Next, all papers marked as (maybe) relevant underwent full-text verification against the criteria, resulting in a total of 7 relevant papers.

\paragraph{Back- and Forward Snowballing}
For snowballing, we used the relevant studies as seed papers, alongside the General Attitudes towards Artificial Intelligence Scale (GAAIS) paper\cite{the_general_attitudes_towards_artificial_intelligence_scale}.
While this paper did not meet our inclusion criteria (general AI instead of LLM-based assistants), we included it due to its methodological relevance to measuring AI attitudes and their relationship with personality. 
For backward snowballing, each papers' references were directly extracted, while forward snowballing used Google Scholar's citation tracking.
All identified studies underwent the same screening procedure as before.
This stepz resulted in additional \textbf{1450} candidate and 5 relevant papers, resulting in 12 papers total.

\subsubsection{Data Extraction}

For each paper in our final sample, we manually extracted general metadata, including the venue, study type (e.g., survey, experiment), and whether the study was SE-specific.
Additionally, we extracted the personality framework used, the dependent variables assessed (i.e., the variables evaluated against personality), and the core findings, i.e., whether personality was associated with these variables, which traits showed such associations in which direction, and supporting quotes for these findings.
Associations were determined based on the authors' own framing of their results, most commonly grounded in statistical testing (e.g., correlation or regression analyses).

\subsection{Review Results and Discussion}

\begin{table*}[!t]
\centering

\caption{Overview of review stuies. Key (not all) findings organized by personality dimensions. Empty trait columns indicate a trait not evaluated on its own. Arrows indicate how a higher presence of a trait correlates with higher ($\uparrow$) or lower ($\downarrow$) outcome values.}
\label{tab:lit-review}
\footnotesize

\renewcommand{\arraystretch}{-4}s

\begin{tabular}{p{0.25cm} p{1.35cm} p{2.6cm} p{1.35cm} p{1.4cm} p{1.4cm} p{1.52cm} p{1.5cm} p{2.8cm}}
\toprule
{} & Framework & Outcome Measures
& \textbf{O} & \textbf{C} & \textbf{E}
& \textbf{A} & \textbf{N} & Other  \\
\midrule

\cite{exploring_peoples_perceptions_of_llm_generated_advice}
& BFI-2\cite{bfifull}
& Usefulness, likeability, follow intention, creepiness of advice
& \notsignificant{}
& \notsignificant{}
& $\downarrow$\,PU
& \makecell[tl]{$\uparrow$\,PU\\$\uparrow$\,Likeability\\$\uparrow$\,Follow\,intention}
& \makecell[tl]{$\downarrow$\,PU\\$\downarrow$\,Advice from\\ \,\,\,\,friend rating}
&
\\

\arrayrulecolor{lightgray}
\cmidrule[0.5pt]{1-9}

\cite{personality_traits_and_the_technology_acceptance_of_chatgpt}
& BFI\cite{bfiadapted} (partial)
& TAM (PEOU, PU, CI)
&
& \makecell[tl]{$\uparrow$\,PU, PEOU\\$\uparrow$\,CI (indirect)}
& $\uparrow$\,PEOU (limited)
&
& \notsignificant{}
&
\\

\cmidrule[0.5pt]{1-9}

\cite{the_influence_of_extraversion_and_neuroticism}
& BFI-44\cite{bfi44} (partial)
& Tech. acceptance, LLM usage, cognitive dissonance
&
&
& \notsignificant{}
&
& \makecell[tl]{$\downarrow$\,Acceptance\\$\uparrow$\,Dissonance\\(small, indirect)}
&
\\

\cmidrule[0.5pt]{1-9}

\cite{who_is_most_likely_to_accept_ai_chatbots}
& BFI-44\cite{bfi44}
& TAM (PU, PEOU, AT, BI)
& $\uparrow$\,PU 
& $\uparrow$\,PU, PEOU
& $\uparrow$\,PEOU
& $\uparrow$\,PU 
& $\downarrow$\,PEOU
&
\\

\cmidrule[0.5pt]{1-9}

\cite{an_analysis_of_students_attitudes_toward_artificial_intelligence}
& CP5F\cite{cp5f}
& Cognitive, affective, behavioral attitude
& \notsignificant{}
& \notsignificant{}
& \makecell[tl]{$\uparrow$\,Open\,attitude\\$\uparrow$\,Behavior}
& \notsignificant{}
& \notsignificant{}
&
\\

\cmidrule[0.5pt]{1-9}

\cite{understanding_university_students_use_of_gen_ai}
& IPIP-50\cite{ipip50}
& GenAI use and perceptions (frequency, purpose, usefulness, preferences)
& $\uparrow$\,Preference of traditional tools
& $\downarrow$\,Usage frequency
& $\uparrow$\,Preference of GenAI
& \makecell[tl]{$\uparrow$\,Ethical\\concerns\\$\downarrow$\,Performance}
& \makecell[tl]{$\uparrow$\,Learning impact\\$\downarrow$\,Concerns}
&
\\

\cmidrule[0.5pt]{1-9}

\cite{personality_correlates_of_academic_use_of_generative_artificial_intelligence_and_its_outcomes}
& BFI-8\cite{bfi8} (partial)
& Academic use of GenAI (frequency focused)
& \notsignificant{}
& $\downarrow$\,Academic use
& 
& 
& \notsignificant{}
&
\\

\cmidrule[0.5pt]{1-9}

\cite{university_students_acceptance_and_usage_of_generative_ai_chatgpt_from_a_psycho_technical_perspective}
& BFI-44\cite{bfi44}
& ChatGPT usage
& $\uparrow$\,Usage
& \notsignificant{}
& \notsignificant{}
& $\downarrow$\,Usage
& $\downarrow$\,Usage
&
\\

\cmidrule[0.5pt]{1-9}

\cite{stimulus_organism_response_framework_of_decision_makers_intention_to_adopt_generative_ai_to_replace_entry_level_jobs}
& Mini-IPIP\cite{miniipip} (partial)
& Trust in GenAI, adoption intention (moderation)
& \notsignificant{}
& $\uparrow$\,Trust--adoption link
& 
& \notsignificant{}
& 
&
\\

\cmidrule[0.5pt]{1-9}

\cite{personality_profiles_and_usage_experience_are_associated_with_trust_and_dependence_on_generative_ai}
& Mini-IPIP\cite{miniipip} (LPA)
& Trust in GenAI
&
&
&
&
&
& ($\uparrow$\,E/A + $\downarrow$\,O/N): ↑ GenAI Trust
($\uparrow$\,N + $\downarrow$\,O/C/E/A): ↓ GenAI Trust
\\

\cmidrule[0.5pt]{1-9}

\cite{personality_and_acceptances_as_preditctors_of_chatgpt_use}
& BFI-10/44, Machiavellianism\cite{bfi10,bfi44,machiavellinism}
& ChatGPT acceptance (e.g., usefulness), usage (e.g., frequency, tasks), human vs AI opinions 
& \notsignificant{}
& \notsignificant{}
& \notsignificant{}
& \notsignificant{}
& \notsignificant{}
& \makecell{Machiavellianism:\\strongest predictor;\\$\uparrow$\,usage and opportunistic\\(undisclosed) use}
\\

\cmidrule[0.5pt]{1-9}

\cite{the_big_five_personality_traits_and_perceptions_of_generative_ai_in_higher_education}
& BFI-44\cite{bfi44}
& Knowledge, attainment, intrinsic/utility value, cost, intention
&
&
&
&
& \makecell[tl]{$\uparrow$\,Cost\\$\downarrow$\,Utility/\\attainment\\value}
& \makecell[tl]{($\uparrow$\,O/A/C): $\uparrow$\,Knowledge,\\cost\\($\downarrow$\,N + $\uparrow$\,O/C): $\uparrow$\,Attain-\\ment, intention; $\downarrow$\,Cost}
\\

\arrayrulecolor{black}
\bottomrule
\end{tabular}

\begin{flushleft}
\small\textit{
BFI\,=\,Big Five Inventory;
IPIP\,=\,International Personality Item Pool;
CP5F\,=\,Five-Factor Personality Questionnaire;
TAM\,=\,Technology Acceptance Model;
PU\,=\,Perceived Usefulness;
PEOU\,=\,Perceived Ease of Use;
CI\,=\,Continuance Intention;
LPA\,=\,Latent Profile Analysis;
O\,=\,Open-Mindedness;
C\,=\,Conscientiousness;
E\,=\,Extraversion;
A\,=\,Agreeableness;
N\,=\,Neuroticism;
n.s.\,=\,Not significant.
}
\end{flushleft}
\end{table*}

\paragraph{Overview}

The review resulted in twelve relevant papers, none of which were found in the selected SE venues, confirming the absence of SE-specific work on this topic. 
The papers span diverse domains (e.g., education\cite{who_is_most_likely_to_accept_ai_chatbots,an_analysis_of_students_attitudes_toward_artificial_intelligence,understanding_university_students_use_of_gen_ai} and psychology\cite{exploring_peoples_perceptions_of_llm_generated_advice,personality_and_acceptances_as_preditctors_of_chatgpt_use,an_analysis_of_students_attitudes_toward_artificial_intelligence}). 
They operationalize personality through the BF Model, though instruments (e.g., BFI-44\cite{who_is_most_likely_to_accept_ai_chatbots,university_students_acceptance_and_usage_of_generative_ai_chatgpt_from_a_psycho_technical_perspective,personality_and_acceptances_as_preditctors_of_chatgpt_use,the_big_five_personality_traits_and_perceptions_of_generative_ai_in_higher_education, the_influence_of_extraversion_and_neuroticism} or Mini-IPIP\cite{personality_profiles_and_usage_experience_are_associated_with_trust_and_dependence_on_generative_ai,stimulus_organism_response_framework_of_decision_makers_intention_to_adopt_generative_ai_to_replace_entry_level_jobs}) and trait labels differ (e.g., Intellect/Imagination vs. Openness\cite{understanding_university_students_use_of_gen_ai}).
One study additionally measured Machiavellianism\cite{personality_and_acceptances_as_preditctors_of_chatgpt_use}.

As summarized in \autoref{tab:lit-review}, the reviewed papers cover a wide range of outcome measures. 
Commonly analyzed is perceived ease of use (PEOU)\cite{personality_traits_and_the_technology_acceptance_of_chatgpt,who_is_most_likely_to_accept_ai_chatbots} and usefulness (PU) of AI tools\cite{exploring_peoples_perceptions_of_llm_generated_advice,personality_traits_and_the_technology_acceptance_of_chatgpt,who_is_most_likely_to_accept_ai_chatbots,personality_and_acceptances_as_preditctors_of_chatgpt_use,understanding_university_students_use_of_gen_ai}, often operationalized through adaptations of the Technology Acceptance Model (TAM)\cite{Davis1989}, a widely used framework, which assumes that ease of use and usefulness jointly determine technology acceptance. 
Beyond TAM-based measures, only usage frequency\cite{personality_and_acceptances_as_preditctors_of_chatgpt_use,understanding_university_students_use_of_gen_ai} and trust in AI tools\cite{personality_profiles_and_usage_experience_are_associated_with_trust_and_dependence_on_generative_ai,stimulus_organism_response_framework_of_decision_makers_intention_to_adopt_generative_ai_to_replace_entry_level_jobs} recur twice.
Each measure is operationalized differently.

Despite the heterogeneity in outcome measures, \insight{almost every paper using a BF personality assessment found at least one BF trait to be associated with measured perceptions}. 
Conscientiousness, Extraversion, and Negative Emotionality each showed individual associations in five studies, though not all papers collected all five traits. 
Effect strength varied, though statistical methods are not directly comparable. 
For example, Wester et al.\cite{exploring_peoples_perceptions_of_llm_generated_advice} reported small predictive power ($R^2$ between $0.11$ and $0.14$), while Glaveanu and Maier\cite{an_analysis_of_students_attitudes_toward_artificial_intelligence} found moderate to strong correlations between Extraversion and attitude toward ChatGPT ($r = 0.56-0.71$). 
Even de Winter et al.\cite{personality_and_acceptances_as_preditctors_of_chatgpt_use}, who found BF trait effects largely disappear after controlling for confounders, still identified the trait Machiavellianism as a significant predictor.
Together, these findings reinforce that personality broadly shapes AI tool perceptions regardless of the framework used for capturing it.

\paragraph{Results by Personality Trait}

Personality was associated with several measures of AI perceptions, though associations varied by trait and outcome. 
\insight{For Conscientiousness, findings point in two directions}. 
Higher Conscientiousness was associated with greater PU and PEOU in two studies\cite{personality_traits_and_the_technology_acceptance_of_chatgpt,who_is_most_likely_to_accept_ai_chatbots} and moderated the positive association between GenAI trust and adoption intention in another\cite{stimulus_organism_response_framework_of_decision_makers_intention_to_adopt_generative_ai_to_replace_entry_level_jobs}.
It was also linked to lower AI usage\cite{understanding_university_students_use_of_gen_ai,personality_correlates_of_academic_use_of_generative_artificial_intelligence_and_its_outcomes}, although usage was operationalized differently across studies. This may suggest that conscientious individuals recognize the usefulness and ease of use of AI tools without necessarily using them more frequently.

\insight{For Extraversion, findings are most consistent concerning PEOU}, where higher Extraversion was associated with higher PEOU in the same two studies as before\cite{personality_traits_and_the_technology_acceptance_of_chatgpt,who_is_most_likely_to_accept_ai_chatbots}, though one effect was limited in magnitude. 
Beyond PEOU, findings were mixed, including associations with more open attitudes and behavior toward AI\cite{an_analysis_of_students_attitudes_toward_artificial_intelligence}, preference for GenAI over traditional tools\cite{understanding_university_students_use_of_gen_ai}, and lower PU\cite{exploring_peoples_perceptions_of_llm_generated_advice}. 
Thus, Extraversion appears consistently related to perceived ease of use, but not to uniformly more positive AI evaluations.

Like Extraversion, \insight{Agreeableness shows mixed findings, though with a more consistent theme}.
Higher Agreeableness was associated with greater PU\cite{exploring_peoples_perceptions_of_llm_generated_advice,who_is_most_likely_to_accept_ai_chatbots}, higher likeability of LLM-generated advice, and greater likelihood of following it\cite{exploring_peoples_perceptions_of_llm_generated_advice}.
However, this receptiveness did not translate further: higher Agreeableness was also associated with lower tool usage\cite{university_students_acceptance_and_usage_of_generative_ai_chatgpt_from_a_psycho_technical_perspective}, greater ethical concerns, and lower perceived performance\cite{understanding_university_students_use_of_gen_ai}. 
As such, agreeable users seem more receptive toward LLM-based tools in direct interactions and simultaneously show more caution about their broader use.

\insight{Negative Emotionality shows a predominantly negative pattern}. 
It was associated with lower PU\cite{exploring_peoples_perceptions_of_llm_generated_advice}, lower PEOU\cite{who_is_most_likely_to_accept_ai_chatbots}, lower technology acceptance and resulting greater cognitive dissonance\cite{the_influence_of_extraversion_and_neuroticism}, and lower usage\cite{university_students_acceptance_and_usage_of_generative_ai_chatgpt_from_a_psycho_technical_perspective}. 
By contrast, one study found it associated with greater perceived learning impact and fewer concerns\cite{understanding_university_students_use_of_gen_ai}, indicating that perceived benefits may coexist with otherwise less favorable perceptions.

Finally, higher Open-Mindedness was associated with greater PU\cite{who_is_most_likely_to_accept_ai_chatbots} and higher usage\cite{personality_and_acceptances_as_preditctors_of_chatgpt_use}. 
At the same time, another study found a preference for traditional tools over GenAI\cite{understanding_university_students_use_of_gen_ai}. 
This suggests that greater openness may facilitate exploration and use of LLM-based tools without necessarily resulting in a preference for them over established alternatives.

A notable finding from two studies is that personality traits do not operate in isolation: specific traits configurations were jointly associated with the respective outcome measures\cite{personality_profiles_and_usage_experience_are_associated_with_trust_and_dependence_on_generative_ai, the_big_five_personality_traits_and_perceptions_of_generative_ai_in_higher_education}. 
As such, some of the observed single-trait associations may partly reflect broader personality configurations, which would not be visible when evaluating each trait independently.

\begin{mybox}{\ques{1}: Personality and AI assistants}
To answer \ques{1}, the literature documents many associations between personality and AI-tool perceptions.
However, these appear outcome-specific rather than creating a uniform mapping between personality dimension and perception. 
The same trait may relate differently to multiple variables, and even across variables, a trait might correlate in one, but not anymore in another study. 
This diversity in itself is not surprising, giving the complexity of personality as a construct.
However, while the literature does not support broad claims, it highlights (1) that personality is associated with how we perceive the AI tools we interact with, even if it cannot provide one generalizable answer, and (2) personality is an important predictor for perceptions, especially regarding usefulness and ease of use, though others might prove to be just as relevant with more studies. 
The notion that such personality impacts extend to SE professionals is not proven yet, but highly probable.
Still, no study has examined this in an SE context, further motivating this work.
\end{mybox}

\section{Empirical Study}

In this section, we address \ques{2} with a survey examining how developers' personality traits are associated with their expectations of and experiences with AI coding assistants. 
Building on the literature review, we explore whether the associations between personality and perceptions of AI tools identified for \ques{1} also emerge among professional software developers. 
As no prior SE-specific evidence exists, the survey is designed to provide first indications of such associations rather than to confirm specific trait effects.

\subsection{Questionnaire}

We conducted a survey with three companies (Feb-Apr 2025, two-week response window) as part of a larger study on developers' AI tool adoption. 
We used four question groups: private AI tool usage and attitude, expectations regarding the AI tool's introduction into their professional work, experiences using the AI tool in everyday work, and the BFI-2-S\cite{SotoJohn2017}.

Given the lack of established instruments for these specific contexts, we developed the survey through multiple feedback cycles with our industry partners. 
Our scoping review revealed heterogeneous personality-perception associations across studies and contexts. 
Accordingly, we structured our survey questions around broad constructs capturing both expectations and lived experiences, rather than narrowing to single outcome measures.
The private AI tool section included one matrix on adoption attitudes and items on which tools participants used, usage frequency, duration, and satisfaction.
The expectations and experiences sections used three matrices on benefits, challenges, and usage areas. It was derived from the 2024 Stack Overflow Developer Survey\cite{stackoverflow2024survey}, supplemented by findings from Mailach et al.\cite{mailach}, and company input. 
Expectations questions asked developers to reflect on initial reactions to the tool introduction, while experience questions paralleled these but captured actual experiences since then.
Approximately one year later, we captured the experiences again, mirroring the previous experience questions.
In this paper, we analyze the private attitude matrix and the first two expectations and experiences matrices from both surveys (benefits and challenges). The questionnaires and responses are available in the supplementary material.

In the first survey, we further used the BFI-2-S\cite{SotoJohn2017}, the 30-item short form of the BF Inventory-2 (BFI-2)\cite{bfifull}.
The instrument assesses the five personality traits (Extraversion, Agreeableness, Conscientiousness, Negative Emotionality, and Open-Mindedness).
We selected BFI-2-S for its psychometric validation (tested in two studies, part-whole $\text{correlations}=.93$\textminus$.97$ and $.86$\textminus$.95$, $\alpha=.73$\textminus$.83$ and $=.39$\textminus$.79$, respectively\cite{SotoJohn2017}), its prevalence in prior research, and compact 30-item format enabling feasibility in on-site deployment.

\subsection{Analysis}

\paragraph{Construct formation}
After data collection, we preprocessed the responses to include only participants that finished the entire survey. 
Then, we used the BFI-2-S\cite{SotoJohn2017} manual, to score each of the five traits per remaining participant.

We then built five conceptually defined composite measures from the responses in the first survey: positive and negative expectations (\expectpos{} and \expectneg{}) when using the AI assistant at work, positive and negative experiences in the same situation (\experiencepos{} and \experienceneg{}), and experiences when privately using the assistant for programming (\experiencepriv{}).  
For simplicity, we refer to them as constructs.
In the second survey, we built three such constructs (\experiencepos{}, \experienceneg{}, and \experiencepriv{}), as expectations were no longer collected. 
Each construct contained 5-7 Likert items, whose responses we rescaled to a 1-5 scale, reverse-coded where necessary, and calculated as a mean score.
These constructs capture broad, conceptually related benefits and challenges. 

To assess whether the grouped items were internally coherent despite this breadth, we evaluated their internal consistency using Cronbach's alpha. 
For constructs with $\alpha < .70$, we additionally inspected corrected item-total correlations using Pearson's $r$, assessing linear covariation between each item and the mean of the remaining construct items.
This allowed us to distinguish lower $\alpha$ caused by misaligned items from those caused by items correlating weakly but in the same direction

\paragraph{Personality--perception associations}
To examine how developers’ personality traits relate to their perceptions, we fitted a separate multiple linear regression for each questionnaire construct, with the construct score as the dependent variable and all five BF traits entered simultaneously as predictors. 
The null hypothesis for each model is that none of the BF traits is associated with the respective score.

We fitted models for all constructs available in each survey and used Jarque--Bera, and Breusch--Pagan tests to assess normality, and heteroscedasticity, respectively.
To distinguish changes over time from changes in sample composition, we additionally re-estimated the Survey 1 models using only the participants who also completed Survey 2.
We further calculated 95\,\% confidence intervals for the regression coefficients and did not adjust for multiple testing across models, given that each regression addresses a distinct construct rather than repeated comparisons of the same outcome.
Instead of interpreting individual coefficients as confirmed effects, we assess whether the observed associations are consistent in direction with those reported in the literature review and whether they recur across survey waves.
As such, these exploratory analyses aimed to investigate associations between personality and developers' perceptions across the two survey waves. 

\paragraph{Item-level follow-up analyses}
To further review the identified associations in the regression models, we followed up with an exploratory item-level analyses for each significant association. 
We calculated Pearson correlations between the associated BF traits and each relevant construct's individual items. 
These analyses were not intended as additional significance tests, but to assess whether the construct-level association was broadly reflected across its items or appeared stronger for particular aspects of the constructs.

\paragraph{AI tool usage in development of analysis scripts}

We used GitHub Copilot to assist with code generation during development of the analysis scripts. 
We maintained full human oversight by individually reviewing all generated code for correctness, security, and alignment with our methodology. 
We manually authored any critical logic and data pipelines and all suggestions were evaluated and often times substantially authored manually. This tool accelerated routine implementation tasks but did not influence research decisions, analytical approaches, or interpretations.

\subsection{Results}
\paragraph{Descriptive information}
The first survey ran from February 14 to April 29, 2025 (\textasciitilde{}2 weeks response window per company), collecting 29 valid responses. 
The second survey was conducted one year later (April 1st to 16th, 2026) in two of the three companies, yielding 11 responses. 
Participation was voluntary, and all participants gave their informed consent prior to the survey.
Demographics were collected only in Survey 1 and linked to Survey 2 via pseudonymized IDs.

Our participants were software developers with diverse backgrounds.
They had general programming experience ranging from six months to twenty years, with professional experience concentrated in the 1-3, 4-6, and >10 year categories (N=8, 8, 9 respectively). 
Most worked in medium-sized (4-7 developers, N=13) or large teams (>7 developers, N=10). 
All participants had officially used their assigned coding tools, primarily ChatGPT and GitHub Copilot (either commercial or private company instances) for three months to two years.

\paragraph{Constructs scores}

\begin{table}
\caption{Internal consistency and averages of constructs in Survey 1 (S1) and Survey 2 (S2): total mean score (Avg), mean change per person between S1 and S2 ($\Delta$), Cronbach's alpha ($\alpha$), and 95\% confidence interval bounds ($CI_L$, $CI_U$).}
\label{tab:internal_consistency_ini2}
\footnotesize
\setlength{\tabcolsep}{4pt}
\begin{tabular}{lrrrrrrrrr}
\toprule
 & \multicolumn{2}{c}{Avg} & $\Delta$ & \multicolumn{2}{c}{$\alpha$} & \multicolumn{2}{c}{$CI_L$} & \multicolumn{2}{c}{$CI_U$} \\
Construct & S1 & S2 &  & S1 & S2 & S1 & S2 & S1 & S2 \\
\midrule
\experiencepriv{} & 4.1 & 4.4 & 0.40 & 0.68 & 0.77 & 0.46 & 0.49 & 0.83 & 0.93 \\
\expectneg{} & 2.8 &  &  & 0.71 &  & 0.51 &  & 0.85 &  \\
\experienceneg{} & 2.6 & 2.6 & 0.10 & 0.51 & 0.55 & 0.17 & -0.08 & 0.75 & 0.86 \\
\expectpos{} & 3.5 &  &  & 0.8 &  & 0.66 &  & 0.89 &  \\
\experiencepos{} & 3.3 & 3.7 & 0.50 & 0.79 & 0.58 & 0.64 & 0.02 & 0.89 & 0.87 \\
\bottomrule
\end{tabular}
\end{table}

Construct scores range from 1 to 5, with higher values indicating stronger agreement with the respective construct; for negative constructs, higher scores thus reflect more negative perceptions. 
As shown in \autoref{tab:internal_consistency_ini2}, \experiencepriv{} had the highest mean scores in both surveys, while the remaining constructs were closer to the scale midpoint. 
Among participants who completed both surveys, within-person changes were minor to modest across the three repeated constructs, ranging from $+.1$ to $+.5$ points.

Half the constructs showed lower internal consistency ($.5 < \alpha < .7$). 
We performed item-level inspection to verify whether this reflected true item misalignment or just limited sample variability and revealed positive but heterogeneous item correlations, suggesting weak but consistent directionality (e.g., \experiencepriv{} with $r=.14$\textendash{}$.69$ in S1).
Full item-level correlations can be found in the supplementary material.
Given the small sample size, we retained the constructs but exercise particular caution interpreting negative experiences in Survey 2, which showed greater item heterogeneity.

\paragraph{Found personality--perception associations}

We ran linear regressions separately for each survey, with results for Survey 1 (S1) shown in the upper part of \autoref{tab:regression_result} and Survey 2 (S2) below. 
While small sample sizes limit our ability to assess all model assumptions conclusively, the Jarque--Bera and Breusch--Pagan tests provided no evidence of violations of residual normality or homoscedasticity.
As outlined before, individual $p$-values are interpreted as exploratory evidence rather than confirmatory hypothesis tests.

Significant predictors are printed in bold. 
In S1, significant coefficients were observed in the \experiencepriv{} and \experienceneg{} models. 
The models explained 20\,\% and 25\,\% of the variance, respectively, though adjusted $R^2$ values were even lower with .02 and .08, respectively.
Neither model was significant overall, providing no evidence that the personality dimensions jointly explain the respective construct.
\insight{No trait was associated with developers’ recalled expectations}.

Extraversion was positively associated with \experiencepriv{} ($\beta=.34$, $p=.04$): a one-point higher Extraversion score was associated with a .34-point more positive private programming experience, holding the other personality dimensions constant. 
Agreeableness was likewise positively associated with \experienceneg{} ($\beta=.58$, $p=.03$), corresponding to a .58-point higher negative-experience score per one-point increase in Agreeableness. 
Thus, within these exploratory models, \insight{higher Agreeableness was associated with more negative reported work-experiences}, while \insight{higher Extraversion was associated with more positive private-use experiences.}

One year later, S2 showed a different pattern of observed personality associations. The \experiencepriv{} model was significant overall ($R^2=.93$, adjusted $R^2=.87$, $p_F=.01$). 
Open-Mindedness ($\beta=.26$, $p=.04$) and, as in S1, Extraversion ($\beta=.46$, $p=.01$) showed significant positive associations with \experiencepriv{}. 
In contrast to S1, \insight{significant coefficients were also observed in the \experiencepos{} model for Agreeableness} ($\beta=.68$, $p=.02$) \insight{and Extraversion} ($\beta=.40$, $p=.03$). The model explained 80\,\% of the outcome variance (61\,\% adjusted), although it did not reach overall significance ($p_F=.07$). 
No significant coefficients were observed for \experienceneg{}. 

Re-estimating the S1 models for the 11 participants who also completed S2 (full results in the supplementary material), Extraversion was again associated with private-use experience ($\beta = .46$, $p = .004$), with the same coefficient as in S2. 
The association between Agreeableness and negative experiences did not appear in this subset ($\beta = -.19$, $p = .82$), and neither Agreeableness nor Extraversion was associated with positive experiences ($\beta = -.21$ and $\beta = -.33$, both n.s.).
\insight{Thus, the S1 association between Agreeableness and \experienceneg{} did not hold among developers who completed both surveys, for whom it was instead associated with positive experiences one year later.}

\begin{table}[!t]
\centering
\caption{Regression results per survey wave: coefficients (Coef.), standard errors ($\mathit{SE}$), t-statistics (t) and p-values (P>|t|). Model fit: (adjusted) $R^2$ ($R^2_A$), F-statistic and p-value ($p_F$), and Jarque--Bera ($p_{JB}$) and Breusch--Pagan ($p_{BP}$) test p-values.}

\setlength\extrarowheight{-3pt}
\setlength{\tabcolsep}{2.5pt}
\label{tab:regression_result}

\footnotesize
\begin{tabular}{p{0.5cm} l r r r r r r r r r r r}
\toprule
\multirow{2}{*}{} &  & \multicolumn{4}{c}{Coefficient estimates} & \multicolumn{7}{c}{Model fit} \\
 &  & Coef. & $\mathit{SE}$ & t & P>|t| & R$^2$ &  $\text{R}^2_A$ & F & $p_F$ & $p_{JB}$ & $p_{BP}$ \\
\cmidrule(lr){3-6}\cmidrule(lr){7-13}
\midrule
\multicolumn{13}{l}{\textbf{Survey 1} (N=29)}\\
\cmidrule[0.1pt]{1-13}
\multirow{6}{*}{\rotatebox[origin=c]{90}{\shortstack[c]{positive\\expectations}}} & c & 2.45 & 1.91 & 1.28 & 0.21 & 0.05 & -0.15 & 0.27 & 0.93 & 0.23 & 0.17 \\
 & O & 0.12 & 0.30 & 0.40 & 0.70 &  &  &  &  &  &  &  \\
 & C & -0.00 & 0.23 & -0.00 & 1.00 &  &  &  &  &  &  &  \\
 & E & 0.18 & 0.21 & 0.82 & 0.42 &  &  &  &  &  &  &  \\
 & A & -0.00 & 0.33 & -0.01 & 0.99 &  &  &  &  &  &  &  \\
 & N & 0.04 & 0.21 & 0.17 & 0.87 &  &  &  &  &  &  &  \\
\midrule
\multirow{6}{*}{\rotatebox[origin=c]{90}{\shortstack[c]{negative\\expectations}}} & c & 0.64 & 1.83 & 0.35 & 0.73 & 0.18 & -0.00 & 0.99 & 0.44 & 0.51 & 0.93 \\
 & O & 0.12 & 0.29 & 0.40 & 0.69 &  &  &  &  &  &  &  \\
 & C & 0.14 & 0.22 & 0.64 & 0.53 &  &  &  &  &  &  &  \\
 & E & -0.19 & 0.20 & -0.94 & 0.36 &  &  &  &  &  &  &  \\
 & A & 0.29 & 0.32 & 0.91 & 0.37 &  &  &  &  &  &  &  \\
 & N & 0.30 & 0.21 & 1.46 & 0.16 &  &  &  &  &  &  &  \\
\midrule
\multirow{6}{*}{\rotatebox[origin=c]{90}{\shortstack[c]{private\\ai experience}}} & c & 2.89 & 1.37 & 2.11 & \textbf{0.05} & 0.20 & 0.02 & 1.14 & 0.37 & 0.70 & 0.80 \\
 & O & -0.11 & 0.21 & -0.51 & 0.62 &  &  &  &  &  &  &  \\
 & C & 0.04 & 0.16 & 0.25 & 0.80 &  &  &  &  &  &  &  \\
 & E & 0.34 & 0.15 & 2.22 & \textbf{0.04} &  &  &  &  &  &  &  \\
 & A & 0.01 & 0.24 & 0.05 & 0.96 &  &  &  &  &  &  &  \\
 & N & 0.12 & 0.15 & 0.76 & 0.45 &  &  &  &  &  &  &  \\
\midrule
\multirow{6}{*}{\rotatebox[origin=c]{90}{\shortstack[c]{positive\\experience}}} & c & 2.82 & 1.76 & 1.61 & 0.12 & 0.09 & -0.11 & 0.46 & 0.80 & 0.51 & 0.36 \\
 & O & 0.35 & 0.28 & 1.28 & 0.21 &  &  &  &  &  &  &  \\
 & C & -0.02 & 0.21 & -0.09 & 0.93 &  &  &  &  &  &  &  \\
 & E & -0.05 & 0.20 & -0.26 & 0.79 &  &  &  &  &  &  &  \\
 & A & -0.01 & 0.31 & -0.04 & 0.97 &  &  &  &  &  &  &  \\
 & N & -0.19 & 0.20 & -0.97 & 0.34 &  &  &  &  &  &  &  \\
\midrule
\multirow{6}{*}{\rotatebox[origin=c]{90}{\shortstack[c]{negative\\experience}}} & c & -1.22 & 1.43 & -0.85 & 0.40 & 0.25 & 0.08 & 1.50 & 0.23 & 0.58 & 0.55 \\
 & O & 0.14 & 0.23 & 0.62 & 0.54 &  &  &  &  &  &  &  \\
 & C & 0.15 & 0.17 & 0.88 & 0.39 &  &  &  &  &  &  &  \\
 & E & 0.08 & 0.16 & 0.48 & 0.64 &  &  &  &  &  &  &  \\
 & A & 0.58 & 0.25 & 2.29 & \textbf{0.03} &  &  &  &  &  &  &  \\
 & N & 0.12 & 0.16 & 0.78 & 0.45 &  &  &  &  &  &  &  \\

\midrule
\multicolumn{13}{l}{\textbf{Survey 2 }(N=11)}\\
\cmidrule[0.1pt]{1-13}
\multirow{6}{*}{\rotatebox[origin=c]{90}{\shortstack[c]{private\\ai experience}}} & c & 1.75 & 0.93 & 1.88 & 0.12 & 0.93 & 0.87 & 14.08 & 0.01 & 0.80 & 0.60 \\
 & O & 0.26 & 0.10 & 2.70 & \textbf{0.04} &  &  &  &  &  &  &  \\
 & C & 0.02 & 0.10 & 0.24 & 0.82 &  &  &  &  &  &  &  \\
 & E & 0.46 & 0.11 & 4.15 & \textbf{0.01} &  &  &  &  &  &  &  \\
 & A & 0.15 & 0.17 & 0.88 & 0.42 &  &  &  &  &  &  &  \\
 & N & -0.17 & 0.09 & -1.96 & 0.11 &  &  &  &  &  &  &  \\
\midrule
\multirow{6}{*}{\rotatebox[origin=c]{90}{\shortstack[c]{positive\\experience}}} & c & 0.42 & 1.16 & 0.37 & 0.73 & 0.80 & 0.61 & 4.08 & 0.07 & 0.62 & 0.61 \\
 & O & -0.18 & 0.12 & -1.51 & 0.19 &  &  &  &  &  &  &  \\
 & C & -0.05 & 0.12 & -0.39 & 0.71 &  &  &  &  &  &  &  \\
 & E & 0.40 & 0.14 & 2.96 & \textbf{0.03} &  &  &  &  &  &  &  \\
 & A & 0.68 & 0.21 & 3.26 & \textbf{0.02} &  &  &  &  &  &  &  \\
 & N & 0.06 & 0.11 & 0.53 & 0.62 &  &  &  &  &  &  &  \\
\midrule
\multirow{6}{*}{\rotatebox[origin=c]{90}{\shortstack[c]{negative\\experience}}} & c & 6.53 & 3.60 & 1.81 & 0.13 & 0.44 & -0.11 & 0.80 & 0.60 & 0.63 & 0.31 \\
 & O & 0.53 & 0.37 & 1.43 & 0.21 &  &  &  &  &  &  &  \\
 & C & 0.24 & 0.38 & 0.63 & 0.55 &  &  &  &  &  &  &  \\
 & E & -0.60 & 0.42 & -1.41 & 0.22 &  &  &  &  &  &  &  \\
 & A & -0.93 & 0.64 & -1.45 & 0.21 &  &  &  &  &  &  &  \\
 & N & -0.49 & 0.33 & -1.47 & 0.20 &  &  &  &  &  &  &  \\
\bottomrule
\end{tabular}
\end{table}

\paragraph{Correlations between construct items and personality traits}

\begin{figure}[!ht]
    \centering
    \includegraphics[width=\linewidth]{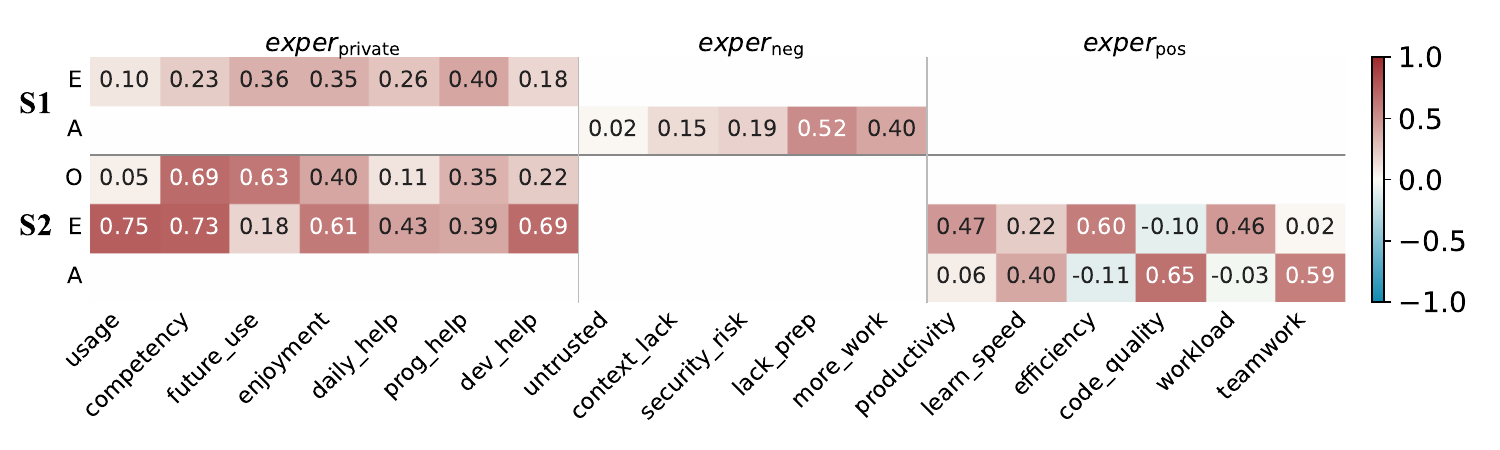}
    \caption{Pearson correlations (r) between individual construct items and personality dimensions that significantly predicted the corresponding construct}
    \label{fig:correlations_significant}
\end{figure}

To examine which items contributed most strongly to construct-level relationships, we inspected item-level correlations (\autoref{fig:correlations_significant}), where each item represents a developer perception. 
For \experiencepriv{}, \insight{Extraversion showed stronger associations in S2 than in S1, particularly for private programming use ($r = .75$) and competency ($r = .73$)}. 
Complementary, Open-Mindedness was most strongly associated with competency ($r = .69$) and future use intention ($r = .63$). 
For \experienceneg{} in S1, \insight{Agreeableness was primarily associated with feeling insufficiently prepared ($r = .52$)}. 
In S2, Agreeableness was instead associated with positive experiences, with strongest associations for code quality ($r = .65$) and teamwork ($r = .59$), while Extraversion related to efficiency ($r = .60$) and productivity ($r = .47$).

\subsection{Discussion}

Our empirical findings provide first indications that the personality--perception associations identified in the scoping review also extend to Software Engineering, though not uniformly across traits or outcomes.
Across the two survey waves, Extraversion showed the most consistent association.
Higher Extraversion was associated with more positive private-use experiences in both waves, as well as in the analysis limited to returning participants.
This aligns with our review findings linking Extraversion to more open attitudes toward AI tools\cite{an_analysis_of_students_attitudes_toward_artificial_intelligence} and preferences for GenAI over traditional tools\cite{understanding_university_students_use_of_gen_ai}.
In S2, Extraversion was additionally associated with positive professional experiences.
Agreeableness showed a less stable pattern: it was associated with negative professional experiences in S1, primarily concerning insufficient preparation, but this association disappeared when restricting S1 to returning participants.
One year later, Agreeableness was instead associated with positive professional experiences.
So, while Extraversion showed a recurring association, differences for Agreeableness may reflect changes in sample composition and developers' experiences over time.

Beyond these temporal differences, the item-level analyses revealed outcome-specific associations.
Professionally, Extraversion related mainly to perceived efficiency and productivity, whereas Agreeableness relates to code quality and teamwork.
Open-Mindedness, which was associated with positive private-use experiences in S2, relates particularly to perceived competency and future use intention.
These findings mirror the scoping review, where personality traits were similarly associated with different aspects of AI-tool perception rather than uniformly positive or negative evaluations.
Further, the review also reported mixed findings for Agreeableness and associations between Open-Mindedness and AI-tool use.
By contrast, the associations for Negative Emotionality is absent in our developer sample.
This discrepancy may reflect differences in population and context, although the small sample also limits our ability to detect associations.
A reasonable explanation for the differences between expectations, professional experiences, and private use is the influence of organizational context.
Institutional constraints may shape developers' initial expectations and workplace interactions, which do not exist in a private setting.
Similarly, personality-related differences might become more apparent as developers gain experience with AI tools.
However, organizational influence was not measured directly, and the different associations across waves do not establish a temporal effect. So, we refer these hypotheses to be evaluated for future research.

\begin{mybox}{\ques{1} and \ques{2}: Towards a Theory}
    Across both studies, personality emerges as a  plausible individual-difference variable worth considering in human-centered AI-for-SE research.
    As a first step towards a theory, our findings suggest three tentative propositions to be tested in future work: (i) personality relates to developers' actual experiences with AI tools rather than to their expectations, (ii) trait effects are outcome-specific, with different traits relating to different perceived benefits, and (iii) these effects depend on context, potentially emerging more clearly with growing tool experience and less institutional influence.

    For \textit{organizations}, this suggests that support beyond initial onboarding may benefit developers more than uniform, one-time rollouts. 
    For \textit{tool builders}, personality-aware designs could be worth investigating, as different developers might benefit from different configurations, e.g., through personalization options or adaptive interfaces.
    For \textit{developers}, our findings suggest exploring the tool configurations already available to match their own working preferences instead of using default setups as they are.
\end{mybox}

\section{Threats to Validity}
\paragraph{Review Threats}
As the review was conducted primarily by the first author, selection and interpretation bias are possible. 
To mitigate this, the second author independently re-screened the first 20 relevant titles of the manual search and all relevant titles from the SE venues, confirming the decisions made. 
We complemented the result cutoff in the database search and the focus on six leading venues with snowballing and AI-assisted searches to broaden coverage and mitigate the risk of missing papers.
All AI-suggested papers were manually verified and screened, to limit additional threats. 
Including only peer-reviewed studies (\excr{E4}) that measure personality with validated instruments (\incr{S5}) further ensures a quality baseline across the included evidence. 
Finally, since we synthesize reported results, publication bias may favor significant findings.
We thus read the review as showing that personality--perception associations occur, not how strongly.

\paragraph{Empirical Study Threats}
Although the developer study is intended only as an initial exploration of whether the findings from the literature review also hold in SE, we addressed potential threats to validity as follows.
The most prevalent threat to our conclusion validity is our sample size (29 Wave 1, 11 Wave 2), which limits statistical power and the stability of estimates with five simultaneous predictors, and may inflate model fit in Wave 2. 
This reflects practical constraints of our study setup: we recruited only professional developers from three industrial companies rather than relying on convenience sampling (e.g., university students), and recruiting time-constrained practitioners for a \textasciitilde{}35-minute survey is challenging. 
The smaller follow-up sample in the second wave resulted from organizational changes, including the loss of one company, so differences between waves may partly reflect sample composition. 
We addressed this by re-estimating the Survey 1 models for the retained participants. 
Nevertheless, we recognize these risks and emphasize throughout the paper that our analysis is exploratory: our conclusions are a stepping stone for future studies to build upon and verify, not confirmatory evidence.

Using self-reported perceptions is appropriate in our case despite potential self-report bias because personality-perception associations are inherently subjective constructs.
Voluntary participation may additionally introduce self-selection bias.
While our company contacts encouraged developers to participate regardless of how 
positive or negative their experiences were, we also do not rely on interpreting how positive or negative these perceptions actually are.
Still, if participation depended on developers' views of AI tools, the restricted range of experiences would more likely weaken than inflate the observed associations.

\section{Conclusion}

Developers' personalities may shape their perceptions of LLM-based coding assistants, yet this relationship remains largely unexplored. A scoping review identified twelve studies linking personality to AI-tool perceptions, particularly usefulness and ease of use, with varying associations across traits and outcomes, but none focused on developers.

To address this gap, we conducted an exploratory two-wave survey across three companies (29 and 11 respondents). No personality dimension was associated with recalled expectations, but associations emerged for actual experiences, particularly in private programming. Higher Extraversion was consistently associated with more positive private-use experiences across both waves, whereas Agreeableness showed negative associations in the first wave, positive in the second.

Consistent with the review, findings suggest context- and outcome-dependent associations rather than stable trait-specific effects. Our work offers initial empirical grounding for a theory of developer personality and AI-tool perception, deriving propositions for future research with larger samples, prospective expectation measures, and more fine-grained assessments of AI-tool experiences.

\section*{Funding}
This work was supported by the Saxon State Ministry for Science, Culture and Tourism (SMWK) through the “Center for Scalable Data Analytics and Artificial Intelligence Dresden/Leipzig” (ScaDS.AI), and by the European Social Fund (ESF) together with the German state of Saxony under grant number A.100760692 (Project: Teaching-AI).

\bibliographystyle{IEEEtran}
\bibliography{references,litrev,frameworks}

\end{document}